\documentclass[letterpaper]{article} 
\usepackage{aaai23}  
\usepackage{times}  
\usepackage{helvet}  
\usepackage{courier}  
\usepackage[hyphens]{url}  
\usepackage{graphicx} 
\usepackage{natbib}  
\usepackage{caption} 
\usepackage{algorithm}
\usepackage{algorithmic}

\usepackage{newfloat}
\usepackage{listings}
\DeclareCaptionStyle{ruled}{labelfont=normalfont,labelsep=colon,strut=off} 
\floatstyle{ruled}
\newfloat{listing}{tb}{lst}{}
\floatname{listing}{Listing}
\usepackage{amssymb}  
\usepackage{amsmath}  
\usepackage{bbm} 
\usepackage{booktabs} 
\usepackage{multirow} 
\usepackage{xcolor}         

\newcommand{\whh}[1]{\textcolor{black}{#1}}

\title{Calibrated Teacher for Sparsely Annotated Object Detection}
\author{
    Haohan Wang\textsuperscript{\rm 1}\equalcontrib, 
    Liang Liu\textsuperscript{\rm 2}\equalcontrib$^{\dag}$, 
    Boshen Zhang \textsuperscript{\rm 2}, 
    Jiangning Zhang \textsuperscript{\rm 2}, 
    Wuhao Zhang \textsuperscript{\rm 2}, \\
    Zhenye Gan \textsuperscript{\rm 2}, 
    Yabiao Wang \textsuperscript{\rm 2}$^{\dag}$, 
    Chengjie Wang\textsuperscript{\rm 2}\textsuperscript{\rm 3}$^{\dag}$, 
    Haoqian Wang \textsuperscript{\rm 1}\thanks{Corresponding authors.}
}
\affiliations {
    \textsuperscript{\rm 1} Shenzhen International Graduate School, Tsinghua University\\
    \textsuperscript{\rm 2} Tencent Youtu Lab\\
    \textsuperscript{\rm 3} Shanghai Jiao Tong University\\
    wang-hh20@mails.tsinghua.edu.cn, \{leoneliu, boshenzhang, vtzhang, wuhaozhang, wingzygan, caseywang, jasoncjwang\}@tencent.com,
    wanghaoqian@tsinghua.edu.cn
}

\usepackage{bibentry}

\begin{document}

\maketitle

\begin{abstract}
Fully supervised object detection requires training images in which all instances are annotated. This is actually impractical due to the high labor and time costs and the unavoidable missing annotations. As a result, the incomplete annotation in each image could provide misleading supervision and harm the training. Recent works on sparsely annotated object detection alleviate this problem by generating pseudo labels for the missing annotations. Such a mechanism is sensitive to the threshold of the pseudo label score. However, the effective threshold is different in different training stages and among different object detectors. Therefore, the current methods with fixed thresholds have sub-optimal performance, and are difficult to be applied to other detectors. In order to resolve this obstacle, we propose a \textit{Calibrated Teacher}, of which the confidence estimation of the prediction is well calibrated to match its real precision. In this way, different detectors in different training stages would share a similar distribution of the output confidence, so that multiple detectors could share the same fixed threshold and achieve better performance. Furthermore, we present a simple but effective \textit{Focal IoU Weight} (FIoU) for the classification loss. FIoU aims at reducing the loss weight of false negative samples caused by the missing annotation, and thus works as the complement of the teacher-student paradigm. Extensive experiments show that our methods set new state-of-the-art under all different sparse settings in COCO. \whh{\textit{Code will be available at \url{https://github.com/Whileherham/CalibratedTeacher}}}.
\end{abstract}

\section{Introduction}
Remarkable progress in fully supervised object detection has been witnessed in recent years. In order to train such an object detector, a large number of training images are required, and bounding boxes in each image are supposed to be exhaustively annotated. However, it could be practically infeasible due to the high labor and time costs with the volume of datasets increasing, and some of the annotations would be unavoidably missing due to human error.

Sparsely annotated object detection (SAOD) aims to handle such a task that each training image may miss some box annotations. It is highly relevant to a popular topic called semi-supervised object detection (SSOD), whose training data consists of a fully labeled part and an unlabeled part. 
It is worth mentioning that these two tasks are complementary
, since even in the semi-supervised settings, the case of missing labels may still appear in the part of labeled data.

\begin{figure}[t]
	\centering
	\includegraphics[width= \columnwidth]{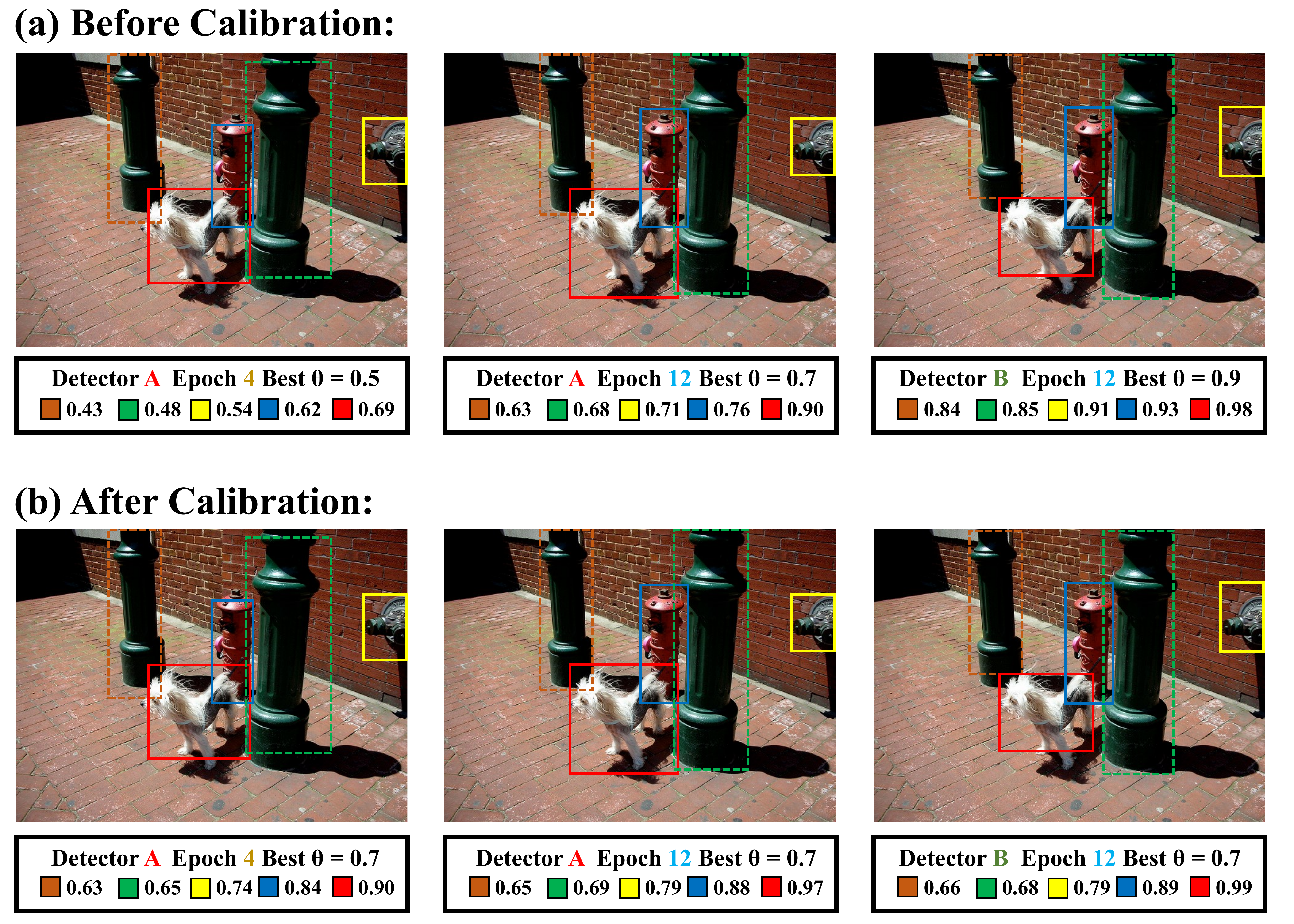}
	\caption{An example of confidence calibration. In order to filter out two \textit{green pillars} which not belong to the detection classes, while keep the \textit{dog} and two \textit{fire hydrants} as pseudo labels, different score thresholds are needed. After calibration, different training stages and detectors are able to share the same score threshold. We show the confidence scores below the image with a color indicator. The dash or solid line indicates the score below or above the threshold.}
	\label{fig:saod}
\end{figure}

\begin{figure}[t]
	\centering
	\includegraphics[width=\columnwidth]{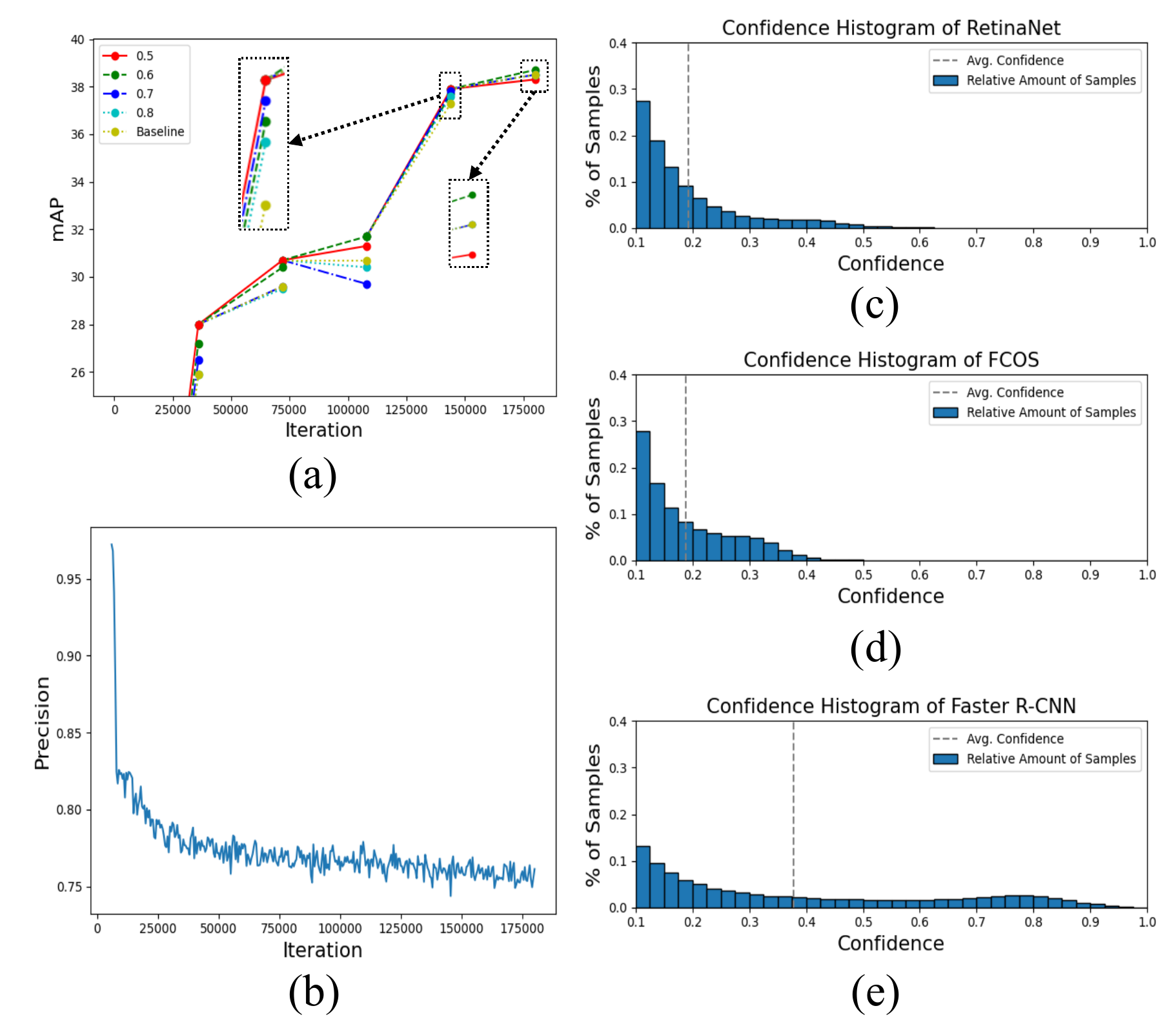}
	\caption{The meaning of confidence score varies in different training stages and among different detectors. (a) compares the effect of different score thresholds in different training stages. (b) shows the precision of pseudo labels during the training with a fixed score threshold. (c)$\sim$(e) demonstrates the score distribution of three models respectively. 
 }
	\label{fig:analysis}
\end{figure}

Considering the relevance of these two tasks, we start from one of the most widely studied methods called ``pseudo labeling" in semi-supervised learning~\cite{tarvainen2017mean}.
Pseudo labeling utilizes the predictions of a teacher network to produce supervision signals over unlabeled data for a student network. Therefore, it is natural to transfer this idea to the case of sparse annotation. In terms of pseudo labeling, the main difference between the semi-supervised task and sparsely annotated task is that the image in the former is either fully annotated or completely unlabeled, while in the later, it is unknown how well the image is labeled. In order to guarantee the precision of pseudo labels, most of previous methods for the semi-supervised task obtain pseudo labels by taking a high score threshold to filter the detection boxes from teacher network. As for the sparsely annotated task, an additional IoU filter is indispensable to drop the predictions which have high overlap with the sparse annotations. The remain boxes along with the sparse annotations compose the final supervision signal.

Following this paradigm, some recent works have tried to generate pseudo labels for the missing annotation boxes to train the model. However, all of the previous methods~\cite{co-mining,saod} adopt a fixed score threshold, which is usually set to a high value, to guarantee only a few accurate pseudo boxes could survive. Such a high score threshold drops lots of moderately confident predictions, which may contain more valuable information (e.g. tiny objects, with rare classes, partially occluded objects), leading to insufficient learning on the missing label. Therefore, a fixed score threshold is not a good choice.

Nevertheless, it is non-trivial to select a better strategy of score threshold, since we find the meaning of confidence score is changing all the time. Firstly we compare the most effective thresholds in different training stages. Concretely, we train RetinaNet under the teacher-student paradigm with six different thresholds. During training we evaluate these models every 36k iterations, of which the best model is utilized to initialize all the models for the next stage. As shown in Fig. \ref{fig:analysis} (a) that the lower thresholds result in the better performance in the earlier stage, but the advantage is narrowing with the training going on. Finally the relatively high thresholds perform better at the end of the training, while the lowest threshold performs worse than not using pseudo labels. 

Besides, as shown in Fig. \ref{fig:analysis} (b), the precision of candidates decreases gradually, which reflects that a stricter threshold is required to ensure the quality of pseudo labels.

Furthermore, there are numerous factors impacting the meaning of confidence score predictions. Fig. \ref{fig:analysis} (c)$\sim$(e) show the score distribution of RetinaNet~\cite{retinanet}, FCOS~\cite{fcos}, and Faster R-CNN~\cite{fasterrcnn} respectively. A wide variation is exhibited due to the different types of loss function(e.g. focal loss v.s. cross entropy loss), different score formulations (e.g. classification score multiply with centerness in FCOS), different model designs(e.g. single stage v.s. two stages). As shown in Fig. \ref{fig:saod} (a), it turns into an engineering task to search for a proper threshold strategy to filter out the correct pseudo labels.

In this paper, we propose to give a unified and specific meaning for the confidence score to simplify the selection of score threshold in pseudo label generating. Inspired by the idea of confidence calibration \cite{cls-calibration}, We propose to calibrate the predictions of the teacher model during training. After calibration. the confidence score is supposed to reflect the expectation of its precision, which is decoupled from the characters and capabilities of detectors. As shown in Fig. \ref{fig:saod} (b), it becomes easier to set a unified threshold for different training stages and detectors after calibration. 

Besides, in order to guarantee the precision of pseudo labels, it is impossible to cover all possible candidates. Hence there still exist some missing annotations not recalled. To resolve this problem, we propose a \textit{Focal IoU Weight}(FIoU) for the classification loss. Supposing the missing annotations tend to have a low IoU with the existing annotations, FIoU down-weights the negative samples who have low IoU with the existing ground truth, with a mechanism similar to focal loss \cite{retinanet}. Therefore, FIoU could be regarded as the complement of SAOD methods. \par

Our contributions are summarized as follows:
\begin{itemize}
\item
We point out that the score threshold is the key factor in pseudo labeling methods, and analyze the obstacle of the existing methods using fixed score threshold in detail.
\item
We propose a general framework for sparsely annotated object detection called Calibrated Teacher, which transforms the confidence score predictions of the teacher network to fit the real precision, so that the model can adaptively generate proper pseudo labels during training.

\item
We propose Focal IoU Weight for classification loss to reduce the negative influence of missing annotations,  which are not mined as pseudo labels.

\item
Extensive experiments show that our methods not only simplify the threshold tuning for different detectors, but also set new state-of-the-art in sparse settings of COCO.

\end{itemize}

\begin{figure*}[t]
	\centering
	\includegraphics[width=0.98\textwidth]{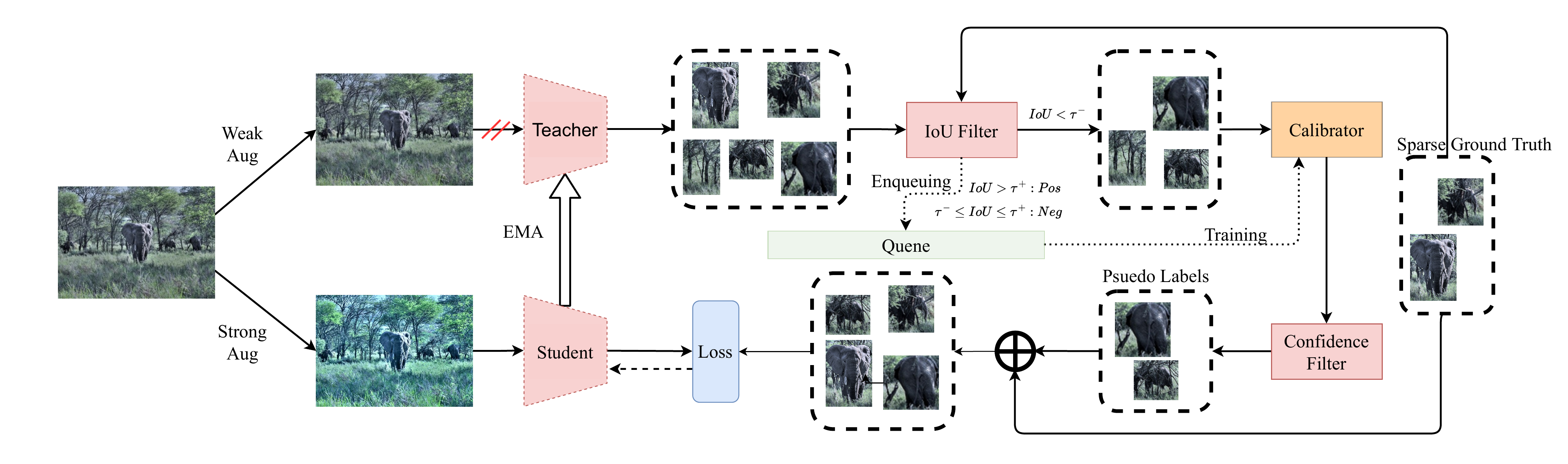}
	\caption{The overall pipeline of our framework. The solid and dotted arrows represent the data and gradient flow, respectively.}
	\label{fig:pipeline}
\end{figure*}

\section{Related work}
\paragraph{Object Detection} is one of the most fundamental problems in computer vision. Various architectures have been proposed such as one-stage methods~\cite{yolov3,ssd} and two-stage methods~\cite{fasterrcnn}. Besides, anchor-free methods draw attention gradually, which usually predict the bounding boxes based on the center points~\cite{centernet}, pseudo center points \cite{fcos} or corner points \cite{cornernet}. with the development of transformer, query-based methods~\cite{detr,deformable-detr} are proposed to model the detection as a set prediction problem. Meanwhile, there exist numerous works on the design of loss function to focus on the class imbalance \cite{retinanet}, the scale variation \cite{iou,giou} and uncertainty measurement \cite{qfl}. However, numerous architectures and training strategies bring great difficulty to design a unified method for sparsely annotated object detection. In contrast, we propose a model-agnostic framework with calibrated confidence predictions. \par

\paragraph{Sparsely Annotated Object Detection} aims at training detectors in a more practical setting, where every training image is likely to contain unannotated instances. Earlier works are usually concentrated on reducing the influence of false negative samples, which should have been assigned to the unannotated ground truth. \cite{saod-softsampling, saod-cvpr} down-weight the negative samples according to the IoU with existing ground truth, or the foreground probability predicted by a pre-trained detector. \cite{saod-icassp} treat the hard negative samples as easy positive ones to avoid the large error signals. However, they could only reach the performance of models trained with partial annotations at best. Recently, \cite{saod-pu} estimate the loss of anchors assigned to missing annotations by that of other positive anchors, but such rough estimation is far from the real loss and thus leads to the sub-optimal performance. \cite{co-mining,saod} adopt a two-stream detection framework to generate the pseudo labels for the missing annotation and improve the performance greatly. However, the performance is sensitive to the threshold of pseudo label selection, which brings difficulty for practical applications. On the contrary, our method provides a general and adaptive strategy to generate pseudo labels. \par

\paragraph{Confidence Calibration for Object Detection} focuses on making the confidence score of detectors consistent with the actual precision. Specifically, for the prediction with the confidence $p$, the expectation of its precision is also supposed to be $p$ if the detector is well-calibrated. 
Although confidence calibration is rarely explored in object detection \cite{od-calibration}, the methodology could be referenced from image classification, such as temperature scaling \cite{cls-calibration}, Plat scaling \cite{plat-scaling}, histogram binning \cite{calibration-histogram-binning}, Bayesian binning \cite{calibration-bayesian-binning}, etc. However, all of these methods require an extra fully annotated validation set to optimize the calibrator after training. On the contrary, we optimize the calibrator with the sparsely annotated training set only, which releases the dependence on the extra clean data and makes it possible to update the calibrator during training. To our best knowledge, this is the first time to introduce the idea of confidence calibration into the sparsely annotated object detection.\par

\section{Method}

\subsection{Overview}

We propose a model agnostic framework \textit{Calibrated Teacher} for sparsely annotated object detection. The overall pipeline is illustrated in Fig. \ref{fig:pipeline}. For clarity, we adopt RetinaNet \cite{retinanet} as the teacher and student detector if not otherwise specified. Notably, other detectors such as FCOS \cite{fcos} and Faster R-CNN \cite{fasterrcnn} could be replaced into our framework without any other modification. 

Following the classical teacher-student paradigm~\cite{tarvainen2017mean}, given the input image $I$, the weakly augmented image and strongly augmented image are taken as the input of the teacher network and student network, respectively. The teacher network is updated by exponential moving average of student network, and the student network is updated in the fully supervised manner but with sparse ground truth and pseudo labels. 

In order to mine the missing labels for the sparse ground truth annotations $\mathbb{G}$, the predictions of the teacher network $\mathbb{P}_t$ are split into two parts: one with low IoU with $\mathbb{G}$, which will serve as pseudo label candidates, and the other with high IoU, which will be pushed into a queue $\mathbb{Q}$ with length $L$. After that, the confidence scores of candidates are transformed by the calibrator $\phi_c(\cdot |\theta_c)$, and will be further filtered by the score threshold $\tau_{s}$ to generate the final pseudo labels. Meanwhile, for every $T$ iteration, the weights of the calibrator are trained with $\mathbb{Q}$. The final pseudo labels along with the sparse ground truth compose the target of the student network.\par

\subsection{Dynamic Threshold: A tedious yet effective trick}

As explained above, even for the same detector, the effective score threshold for pseudo label selection is varying during the training. A relatively low threshold could contribute to the faster convergence in the earlier stage, but harm the performance in the later stage. As a result, an intuitive solution is to adopt a dynamic threshold, which grows gradually with the training going on. In this work, the dynamic score threshold $\tau_{s}$ at epoch $e$ is defined as follows:
\begin{equation}
\tau_s = \tau_0 + (1-\tau_0) \cdot \frac{\ln (e -e^-)}{\ln (e^+ - e^-)}
\end{equation}
where $e^+$, $e^-$ are the total epochs and the beginning epoch to adjust the threshold, respectively. $\tau_0$ is the initial threshold, and we set a relatively low $\tau_0$ for better performance. \par

We experimentally demonstrate the effectiveness of dynamic threshold using RetinaNet and FCOS, as shown in Tab. \ref{fcosrcnn}. However, this strategy still faces the challenge that different detectors do not share the hyper-parameters $e^-$ and $\tau_0$, which requires extra effort to tune. Furthermore, it could be found in Tab. \ref{fcosrcnn} that the performance of Faster R-CNN are especially sensitive to the threshold of pseudo label selection. Even a slight change of the threshold might lead to a performance drop over $10\%$. For such detectors, the pre-defined dynamic strategy has to been designed carefully, which extends the cycle of the model designing greatly. \par

\subsection{Calibrated Teacher: Simplify threshold tuning} 
In order to address the shortcomings of the dynamic threshold, we first analyse the mechanism behind this method. An ideal threshold is closely related to the confidence distribution of the teacher network's prediction, which varies in the different training stages and among different detectors. The dynamic strategy tries to fit the variation of this distribution and adjust the threshold accordingly. However, the variation of confidence distribution is hard to parameterized, which brings great difficulty of designing a pre-defined rules of adjusting threshold. \par

Hence, it is unwise to leverage a pre-defined rules to adapt to the confidence distribution changing continuously. Instead, we turn to focus on the adjustment of confidence distribution and leave the threshold unchanged. This is ideally equivalent to dynamic threshold. However, with the fixed threshold, all we need is to calibrate the confidence distribution into some specific form, which is more robust and easier than adapting the threshold to a varying and non-parametric distribution. Furthermore, once the confidence is calibrated, an effective threshold could also work in other cases, \textit{i.e.}, in other training stages or using other detectors.\par


Then the problem turns into what the target confidence distribution looks like and how we transform the original distribution to that. For the first problem, considering that the precision is the most direct and crucial indicator of the quality of pseudo labels, we expect the confidence score to reflect the real precision. This is the same as the target of confidence calibration in object detection \cite{od-calibration}, which defines a calibrated confidence (unbiased confidence) as follows.\par

Supposing the prediction set is $\{\hat{c_i}, \hat{p_i}, \hat{b_i}\}$, where $\hat{c_i}$, $\hat{p_i}$ and $\hat{b_i}$ are the predicted category, the confidence and the bounding box, respectively. Accordingly, the ground truth assigned to each prediction is denoted as $\{c_i, b_i\}$. Then, the prediction is perfectly calibrated to be unbiased if
\begin{equation}
\begin{aligned}
\mathbb{P}(m=1 | \hat{p} = p) = p. 
\end{aligned}
\end{equation}
where $m=1$ denotes a correct match \textit{i.e.,} $\hat{c_i} = c_i$ and $\text{IoU}(\hat{b_i},b_i)>\tau$ ( $\tau$ is the pre-defined threshold), while $m=0$ indicates a mismatch. \par

Therefore, we naturally consider a regression model for confidence calibration. Specifically, we adopt a logistic calibration model $\phi_c(\cdot |\theta_c)$, and minimize the Negative Log Likelihood of $\phi_c(\hat{p_i} |\theta_c)$ to update the model. \par

\subsection{Online Calibration}
In the field of model calibration, the parameters of calibrators are optimized with the validation set at the end of training. However, there exist two extra challenges in our framework. The first challenge is that there is no validation set available. Although we could utilize training set instead, it is actually noisy due to the missing annotation. The calibrator optimized with such dataset would tend to output the low confidence, as it is trained to reduce the confidence of some correct prediction assigned with the missing ground truth. The second challenge is that the meaning of the score prediction is varying during training, and thus we have to calibrate the teacher network timely, which requires the calibrate model is updating during the training stage. \par

In order to tackle the first challenge, we divide the teacher's prediction in the training set into two parts by an IoU filter. Bounding boxes having a high IoU with at least one of the existing ground truth are more likely to be assigned to the sparse annotation, while those having a low IoU with all the existing annotations possibly cover the missing annotation. Theoretically, the statistics of the two parts are similar, therefore we could optimize the calibrator with the sparse annotations and the corresponding predictions (the high IoU parts). \par

\begin{algorithm}[tb]
\caption{Pseudo labels in Calibrated Teacher}
\label{alg:optimization}
\textbf{Input}: Image $I$, Sparse annotations $\mathbb{G} = \{c_j, b_j\}_{j=1}^M$, \\
         Teacher model predictions at $t$ step $\mathbb{P}_t=\{\hat{c}_i, \hat{p}_i, \hat{b}_i\}_{i=1}^N$,\\
          Calibration queue $\mathbb{Q}$ with length $L$, Calibration model $\phi_c$ with weights $\theta_c$, initialized as an identity mapping. \\
\textbf{Parameter}: Normal and stricter IoU thresholds $\tau^-$, $\tau^+$, Score threshold $\tau_s$, Calibration model update interval $T$.\\
\textbf{Output}: Pseudo labels $\mathbb{L}$, Calibration model weights $\theta_c$, Updated queue $\mathbb{Q}$ ;
\begin{algorithmic}[1] 
\STATE Let $\mathbb{L}=\mathbb{G}$.
\FOR{$i = 1,2,3,...,N$}
    \STATE $\text{iou} = \max_j(\text{IoU}(\hat{b}_i, b_j))$ s.t. $c_i=c_j$
    \STATE $\text{score} = \phi_c(\hat{p}_i | \theta_c)$ // Calibrate model prediction
\IF{$\text{iou} < \tau^-$ and $\text{score} > \tau_{s}$}
    \STATE $\mathbb{L} = \mathbb{L} \cup \{(\hat{c}_i, \hat{b}_i)\}$  // Add to pesudo labels 
\ENDIF

\IF{$\text{iou} > \tau^-$}
    \IF{$\text{iou} > \tau^+$}
        \STATE $\mathbb{Q}$.enqueue($(\hat{p}_i, 1)$)
    \ELSE
    \STATE  $\mathbb{Q}$.enqueue($(\hat{p}_i, 0)$)
    \ENDIF
\ENDIF
\ENDFOR
\IF{$t \text{ mod } T = 0$} 
\STATE // Update the weights of calibration model \\
\STATE $\theta_c = arg\min_{\theta_c}(\textit{\whh{NLL}}(\phi_c(\mathbb{Q} |\theta_c)))$   
\ENDIF

\STATE \textbf{return} $\mathbb{L}$,  $\mathbb{Q}$, $\theta_c$.
\end{algorithmic}
\end{algorithm}

For the second challenge, we introduce a queue $\mathbb{Q}$ with length $L$ during training. At each iteration, for the teacher network's predictions with high IoU, $\{\hat{p_i}, m_i\}$ are pushed into $\mathbb{Q}$, while those from the earliest image are popped, so that $\mathbb{Q}$ only contains the training data in the most recent batches. Therefore, $\phi_c(\cdot |\theta_c)$ is updated with elements in $\mathbb{Q}$. For more details, please refer to Algorithm \ref{alg:optimization}.\par

\subsection{Focal IoU Weight}
Although the calibrated teacher is able to generate pseudo labels properly, it is impossible to cover all the missing annotation. On the one hand, even the model trained with the complete annotation could fail to detect all instances in the training set. On the other hand, the high recall of missing annotation could also result in the low precision, since we have to set a low threshold and thus preserve many false positive predictions. Therefore, the missing but not recalled annotation would still make some positive samples regarded as negative samples. These false negative samples have little influence on the two-stage detectors like Faster R-CNN due to the sampling strategy. However, for the one-stage detector like RetinaNet, they are unlikely to be ignored. Furthermore, the false negative samples usually have a relatively high confidence of foreground, they would further harm the training due to the commonly used focal loss. \par

As a result, it is necessary to reduce the influence of false negative samples. Inspired by \cite{ saod-icassp}, we adopt the IoU with the existing ground truth as the criterion for judging negative samples. It is safe to assign a high weight to those negative samples with high IoU with the existing ground truth labels, as they are less likely to match the missing labels, and vice versa. Analogous to the mechanism of focal loss, where high weights are assigned to negative samples with high foreground confidence, we propose a \textit{Focal IoU Weight} (FIoU) for the classification loss. Considering the formulation of focal loss for the negative samples:
\begin{equation}
    \text{FL}(x) = -\alpha_t x^\gamma\log(1-x)
\end{equation}
where $x$ is the confidence of the foreground. $\alpha_t,\gamma$ are hyper-parameters. Similarly, for the given negative sample, the FIoU weight for the classification loss is defined as:
\begin{equation}
    \text{FIoU}(\text{iou}) = w_0 + k\cdot \text{FL}(\text{iou})
\end{equation}
where $w_0$ and $k$ are hyper-parameters, and the $\text{iou}$ is the max IoU between the negative samples and the existing labels and pseudo labels. \par

\section{Experimental Results}
\subsection{Datasets}

Recent SAOD methods \cite{saod-pu,co-mining,saod-icassp,saod} are mainly evaluated on the challenging COCO-2017 dataset \cite{coco}, 
However, these methods adopt different ways to reconstruct the original training set into the sparsely annotated one. 
For comprehensive comparison, we evaluate our method in almost all of the existing sparse settings:

\noindent \textbf{Split-1:} Following \cite{co-mining,saod-icassp,saod}, for each category $c$ in the training set, $p\%$ annotations are deleted randomly, where $p = \{30,50,70\}$. 

\noindent \textbf{Split-2:} Following \cite{saod}, for each category $c$ in the training set, images containing $c$ are firstly selected. Then for each image, all annotations of $c$ are deleted simultaneously with a probability $p\%$, where $p = \{30,50,70\}$. We make sure that each image contains at least one annotation.\par

\noindent \textbf{Split-3:} Following \cite{saod}, we deleted $p\%$ annotations randomly in a class-agnostic fashion, where $p = \{30,50,70\}$. \par

\noindent \textbf{Easy / Hard / Extreme:} These training sets come from \cite{co-mining}. For each image in the Easy split, one annotation is deleted randomly; For each image in the Hard split, half of the annotations are deleted randomly; For each image in the Extreme split, only one annotation is preserved randomly.


\subsection{Implementation Details}
We take experiments on three common detection methods with ImageNet \cite{imagenet} pretrained ResNet101 \cite{resnet} and ResNet50 for the comparison with state-of-the-art and ablation study, respectively. Our models are trained for $180k$ iterations with a total batch size 16. The learning rate is initialized as 0.01 and gradually decreases to 0.001 and 0.0001 at $120k$ and $160k$ iterations. Other hyper-parameters of the architecture and training schedule are consistent with the implementation in \cite{mmdetection}. \par

\begin{table*}[t]
\centering
\resizebox{0.8\textwidth}{!}{
\begin{tabular}{lccccccccc}
\toprule
\multirow{2}{*}{Method} &  \multicolumn{3}{c}{Split-1} & \multicolumn{3}{c}{Split-2} & \multicolumn{3}{c}{Split-3} \\ \cmidrule[\cmidrulewidth](l){2-4} \cmidrule[\cmidrulewidth](l){5-7} \cmidrule[\cmidrulewidth](l){8-10}
 & 30\% & 50\% & 70\% & 30\% & 50\% & 70\% & 30\% & 50\% & 70\% \\ \midrule
Full Annotation  & \multicolumn{9}{c}{41.4} \\
Baseline & 39.3 & 37.5 & 34.1 & 39.1 & 37.4 & 35.5 & 39.2 & 37.6 & 34.3 \\ \midrule
BRL \cite{saod-icassp} & -- & 32.7 & -- & -- & -- & -- & -- & -- & -- \\
Co-mining \cite{co-mining}  & 36.4 & 32.8 & 24.9 & 36.7 & 33.0 & 24.8 & 36.8 & 32.5 & 25.0 \\

Ours + RetinaNet  & \textbf{40.5} & \textbf{39.3} & \textbf{36.7} & \textbf{40.1} & \textbf{38.9} & \textbf{37.4} & \textbf{40.5} & \textbf{39.0} & \textbf{36.6} \\ \midrule
Unbiased Teacher$^\dagger$~\cite{liu2021unbiased} & \whh{32.0} & 31.1 & \whh{27.9} & 36.4 & 32.9 & 31.4 & 36.0& 32.1 & 30.1 \\
SAOD$^\dagger$ \cite{saod} & 38.5 & 36.2 & 33.0 & 40.0 & 37.2 & 35.9 & 39.7 & 37.4 & \textbf{35.9} \\
Ours + Faster R-CNN$^\dagger$ & \textbf{41.0} & \textbf{39.3} & \textbf{35.3}	&	\textbf{40.8}	& \textbf{39.2}	& \textbf{36.5}	& \textbf{41.0} &	\textbf{39.1} &	\textbf{35.9}  \\
\bottomrule
\end{tabular}
}
\caption{Comparison with recent sparsely annotated object detection methods on three splits of COCO dataset. All the methods adopt ResNet 101 with FPN as backbone. $^\dagger$ denotes using Faster R-CNN rather than RetinaNet as the detector.}

\label{split123}
\end{table*}

As for the confidence calibration, $\tau^+, \tau^-, \tau_{s}$ are set to 0.75, 0.6 and 0.7 for all detectors, respectively. However, most of the prediction of the teacher has an especially low confidence, \textit{e.g.}, lower than 0.2. Prediction under such a low confidence could hardly cover a satisfying result. Therefore, we only consider the prediction whose original confidence is higher than 0.4 during training to reduce the cost of memory and computation. We adopt Plat scaling as the form of the calibrator for its simplicity and effectiveness. For the calibrator training, $T$ is set to 500 and $L$ is the number of predictions of 8000 images. For the FIoU, $w_0$ and $k$ are set to 0.5 and 1.5, while the $\alpha_t$ and $\gamma$ stay consistent with focal loss.\par

The strong augmentation used in our framework contains random resize, contrast normalization, histogram equalization, random solarization, color balance, contrast, brightness, sharpness and posterization, while the weak augmentation includes nothing but random flipping.

\begin{table}[t] 
\resizebox{\columnwidth}{!}{
\begin{tabular}{@{}l|cc|ccccc@{}} 
\toprule 
\multirow{2}{*}{Detector} & \multirow{2}{*}{Full} & \multirow{2}{*}{Base} & \multicolumn{3}{c}{Fixed Threshold} & \multirow{2}{*}{DT} & \multirow{2}{*}{CT} \\ 
\cmidrule(lr){4-6} & & & 0.5 & 0.7 & 0.9 & & \\ 
\midrule RetinaNet & 41.4 & 37.5 & 36.9 & 37.4 & 37.5 & \textbf{38.5} & \textbf{38.5} \\ 
FCOS & 42.5 & 37.9 & 38.0 & 38.2 & 38.0 & 38.7 & \textbf{39.0} \\ 
Faster R-CNN & 42.5 & 37.9 & 11.6 & 27.5 & 39.2 & 38.6 & \textbf{39.3} \\ 
\bottomrule 
\end{tabular} 
}

\caption{Generalization of Calibrated Teacher on different detectors. Full and Base are the abbreviations of full annotation and baseline. DT and CT represent dynamic threshold and Calibrated Teacher, respectively.} \label{fcosrcnn}

\end{table}

\subsection{Comparison with State-of-the-art}
In this subsection we compare our methods with state-of-the-art. Tab. \ref{split123} reports the results in Split-1,2,3. It could be seen that our framework with both RetinaNet and Faster R-CNN could set new state-of-the-art in all sparse settings. Specifically, when adopting RetinaNet as the detector, our methods surpass the counterparts by a large margin (12.8 mAP at most and 7.3 mAP on average). Meanwhile, our methods with Faster R-CNN have the superiority of 3.1 mAP at most and 1.6 mAP on average. It is encouraging to find that our framework improves the performance of baseline methods significantly, which is close to the models trained with the complete training set. This inspires us that in order to relieve the burden of annotating, a sparsely annotated dataset could replace the completely annotated one to some extent. Furthermore, different from the recent work \cite{saod}, which could only be applied on two-stage detectors due to the dependence on the architecture modification of RPN, our framework is model-agnostic, which is convenient and flexible for practical application. In conclusion, the strong performance demonstrates the effectiveness of our methods. \par

\subsection{Generalization to Multiple Detectors}

As explained before, one of the advantages of Calibrated Teacher over dynamic threshold is that different detectors are able to share the same threshold. In order to test this conjecture, we evaluate Calibrated Teacher and dynamic threshold with another two detectors FCOS and Faster R-CNN in Split-1 with $50\%$ missing annotations. For dynamic threshold, we first train models with different fixed thresholds, then the threshold with best performance would serve as $\tau_0$ for better performance. We set $e^-$ to 18 for all three detectors to keep the same dynamic strategy. For Calibrated Teacher, we set the same $\tau_{s}$ as RetinaNet.\par

\begin{figure*}[t]
	\centering
	\includegraphics[width=\textwidth]{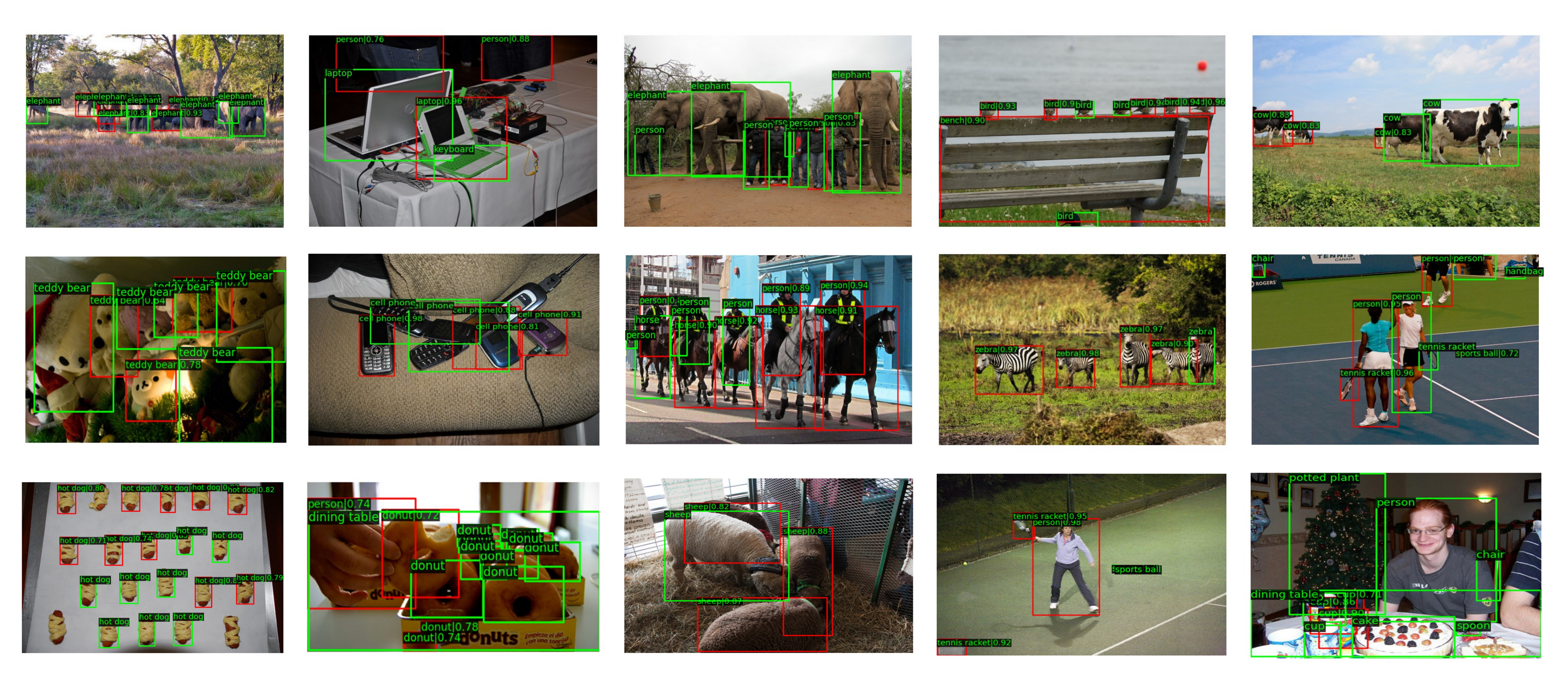} 

	\caption{Qualitative visualization for the results of sparsely annotated object detection. \whh{The 1st and 2nd rows visualize the pseudo labels generated during training, where the green boxes correspond to the available ground truth, and the red boxes indicate the mined pseudo labels. The 3rd row visualizes some failure while mining pseudo labels, where the green and red boxes are available ground truth and pseudo labels.}}

	\label{fig:vis}
\end{figure*}

Tab. \ref{fcosrcnn} summarizes the comparison. It could be found that the dynamic threshold does not always outperform the fixed threshold. This might result from the difficulty of designing a proper dynamic strategy, as different detectors may not share the same effective strategy due to the different confidence distribution. However, Calibrated Teacher is able to surpass both dynamic threshold and fixed threshold in all three detectors, even without adjusting hyper-parameters according to the property of them. Hence we could conclude that Calibrated Teacher is a general framework with better performance. It simplifies the design of hyper-parameters by providing an adaptive strategy to select proper pseudo labels in different training stages, and making it possible that different detectors could share the same hyper-parameters. \par

\begin{table}[t]
\centering
\resizebox{0.75\columnwidth}{!}{
\begin{tabular}{@{}ccccc@{}}
\toprule
\begin{tabular}[c]{@{}c@{}}Calibrated \\ Teacher\end{tabular} & Focal IoU & Easy & Hard & Extreme \\ \midrule
 &  & 37.1 & 35.3 & 26.7 \\
$\checkmark$ &  & 37.7 & 35.5 & 27.4 \\
 & $\checkmark$ & 37.7 & 35.6 & 27.8 \\
$\checkmark$ & $\checkmark$ & \textbf{38.1} & \textbf{36.4} & \textbf{28.7} \\ \bottomrule
\end{tabular}
}

\caption{Ablation study on Easy, Hard and Extreme sets}

\label{split456}
\end{table}

\subsection{Ablation Study}
\subsubsection{Effectiveness of Each Module}
We further validate the effectiveness of each module in Easy, Hard and Extreme sets. As shown in Tab. \ref{split456}, both Calibrated Teacher and Focal IoU outperform the baseline in all three settings when used alone, with the superiority of 0.7 mAP and 1.1 mAP at most, respectively. Furthermore, when combined these two modules, the performance gain is greater than the sum of the respective gains. Taking experiments in Hard set as an example, Calibrated Teacher and Focal IoU could only bring the improvement of 0.2 mAP and 0.3 mAP, but the combination could result in a performance gain of 1.1 mAP. This is because the two modules are complementary. Focal IoU down-weights the false negative anchors to avoid great misleading supervision, which is helpful to train a better model to generate better pseudo labels. \par

\begin{table}[]
\centering
\resizebox{\columnwidth}{!}{
\begin{tabular}{@{}lcccccc@{}}
\toprule
$\tau_{s}$ & 1 & 0.55 & 0.65 & 0.7 & 0.75 & 0.85 \\ \midrule
\whh{mAP/$AP_{50}$} & 37.5/57.3 & 37.6/56.4 & 38.3/57.2 & \textbf{38.5}/57.2 & 38.1/56.9 & 38.0/\textbf{57.3} \\ \bottomrule
\end{tabular}
}

\caption{Sensitivity of $\tau_{s}$ for the score filter after calibration.}

\label{theta}
\end{table}

\begin{table}[]
\centering
\resizebox{\columnwidth}{!}{
\begin{tabular}{@{}lccccc@{}}
\toprule
$w_0$ & 0.5 & 0.5 & 0.5 & 0 & 1 \\ 
$k$ & 1.0 & 1.5 & 2.0 & 1.5 & 1.5 \\ \midrule
\whh{mAP/$AP_{50}$} & 39.2/58.2 & \textbf{39.3}/58.1 & \textbf{39.3}/\textbf{58.3} & 33.9/49.0 & \textbf{39.3}/58.1  \\ \bottomrule
\end{tabular}
}
\caption{Sensitivity of hyper-parameters in Focal IoU.}
\label{k}
\end{table}

\subsubsection{Impact of Hyper-parameters}
In this subsection we first compare the Calibrated Teacher with different $\tau_{s}$ in Split-1 with $50\%$ missing annotations. It could be found in Tab. \ref{theta} that all five $\tau_{s}$ could outperform the baseline, which demonstrates the superiority over using a fixed threshold. Moreover, we set $\tau_{s} = 0.7$ due to its best performance. \par

\begin{table}[]
\centering
\resizebox{0.7\columnwidth}{!}{
\begin{tabular}{@{}lccc@{}}
\toprule
L & 4000 & 8000 & 16000 \\ \midrule
mAP/$AP_{50}$ & \textbf{33.6}/\textbf{51.8} & 33.0/50.7 & 33.0/50.8 \\ \bottomrule
\end{tabular}
}
\caption{Comparison in the queue size $L$. }
\label{queuesze}
\end{table}

\begin{table}[]
\centering
\resizebox{0.7\columnwidth}{!}{
\begin{tabular}{@{}lccc@{}}
\toprule
T & 250 & 500 & 1000 \\ \midrule
mAP/$AP_{50}$ & 33.0/50.6 & 33.0/50.7 & \textbf{33.2}/\textbf{51.0} \\ \bottomrule
\end{tabular}
}
\caption{Comparison in the training interval $T$.}
\label{T}
\end{table}
Then we evaluate Focal IoU with different $w_0$ and $k$ in Split-1 with $50\%$ missing annotations. Tab. \ref{k} reports the results that Focal IoU is more robust to $k$ than $w_0$. When $w_0 = 0$, there exists a great drop of performance. This is because most negative samples have low IoU with ground truth bounding boxes, and of which most are true negative samples. These samples contribute a lot to the classification loss, but their weights tend to be zero when $w_0$ is very small. Therefore, we need a trade-off to balance down-weighting the false negative samples and saving the true negative samples. Hence we use $w_0 = 0.5$ and $k=1.5$ in this work. \par

\whh{Next we analyze the effect of the queue size. Concretely, we train the models for $90k$ iterations in Hard set, and decrease the learning rate at $60k$ and $80k$ iterations. As shown in Tab. \ref{queuesze}, when the queue size $L$ is the number of predictions of 4000 images, the model performs best. This is because a smaller $L$ means that the data stored in the queue comes from more recent iterations and thus has a more similar distribution to the current data. Therefore, if $L$ is not too small to serve as a training set, a relatively small $L$ could result in better performance. }

\whh{Finally we experiment with different intervals at which the calibrator is trained in Hard set. Specifically the model is trained for $90k$ iterations and $L$ is set 8000. Tab. \ref{T} reports the results, and it turns out that our method is robust to the interval $T$.}

\subsubsection{Qualitative Results}

\whh{The first two rows of} Fig. \ref{fig:vis} show the qualitative results of the pseudo labels generated. Our methods are able to mine the small or occluded objects in dense scenes, which tend to miss in annotating. For other more obvious objects (like the zebras in the second row), our methods could provide pseudo labels with extremely high quality, releasing the burden of manual working. \par

\begin{figure}[t]
	\centering
	\includegraphics[width=.95\columnwidth]{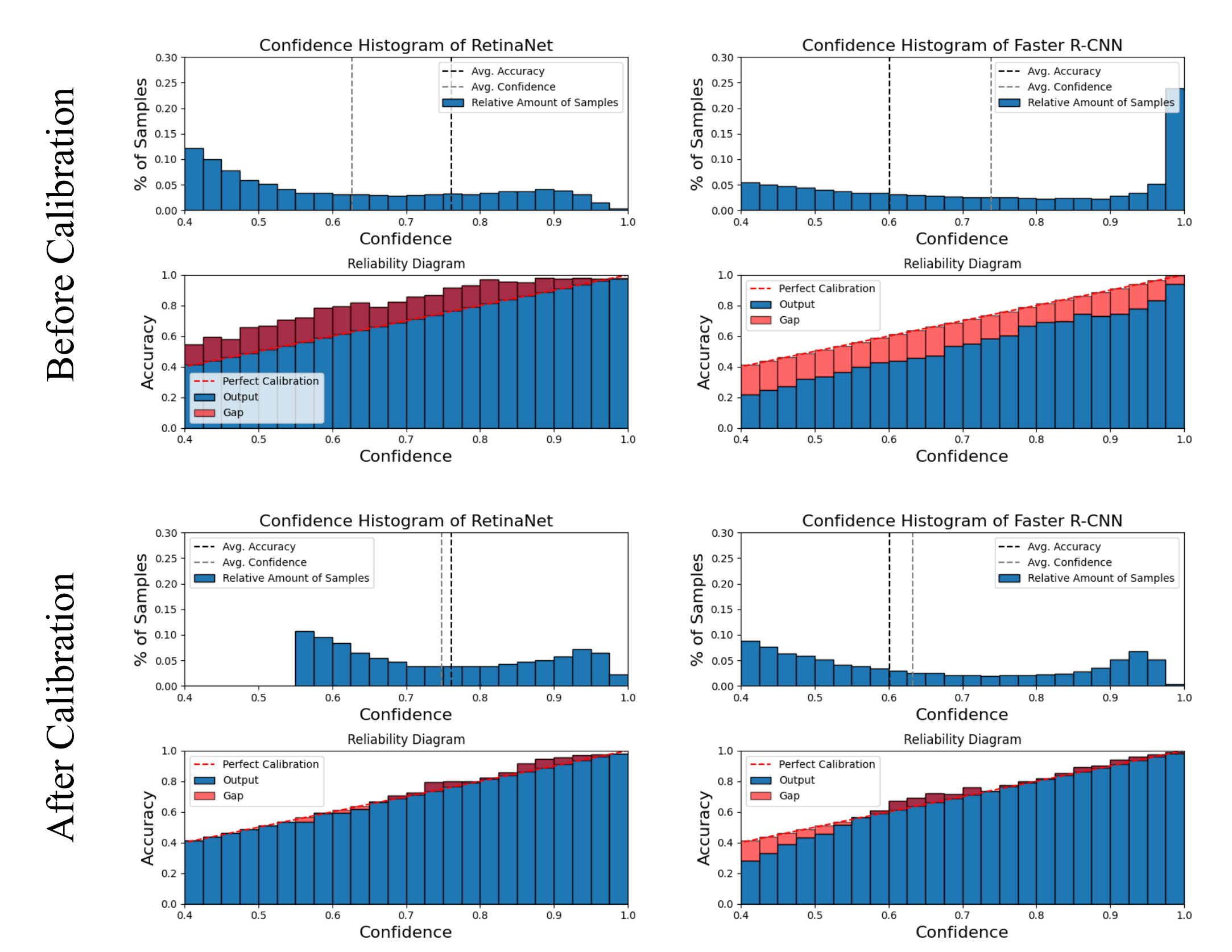}

	\caption{Comparison on the score distribution and ECE between original and calibrated teacher's prediction during training. The first and second columns are the results in the last 500 iterations of RetinaNet and Faster R-CNN. For each sub-figure, the first row shows the score distribution, while the second row illustrates the precision of the pseudo label candidates in each score interval, where the area of red and orange regions represent the degree of the under-confidence and over-confidence, respectively.}
	\label{fig:ece}
\end{figure}

Moreover we visualize the qualitative effect of confidence calibration. Concretely we compare the score distribution and Expected Calibration Error (ECE) \cite{calibration-bayesian-binning} of the pseudo label candidates of RetinaNet and Faster R-CNN in Fig. \ref{fig:ece}. It could be found that after calibration the confidence score is closer to the precision in each interval, and the ECE of the prediction is lower. It indicates that our methods calibrate score distribution into a unified and meaningful form, so that one can design the score threshold more easily and effectively.\par

\subsection{Failed Cases}
\whh{
In the 3rd row of Fig. \ref{fig:vis}, we further show some failed cases when generating the pseudo labels and analyse the possible reasons. In the 1st column we miss three hot dogs while recalling the others. Since the lowest calibrated confidence of the recalled hot dogs is close to the $\tau_d$ (0.71 v.s. 0.7), the missing hot dogs are very likely to be refused by the confidence filter. In the 2nd column we miss one instance in a pile of donuts. This is possibly because the missing donuts are too close to be distinguished from its neighbors, making its corresponding prediction refused by the IoU filter. In the 3rd column the person behind the mesh wire is not recalled, which might result from the great variation in color. In the 4th column we fail to find the ball in the left mostly due to its extremely tiny volume. Finally the right person is excluded from the pseudo labels, which is possibly because only an arm is visible in the picture. Although our method fails to recall some instances, it is actually unavoidable when guarantee the precision of pseudo labels, which is why we need the Focal IoU Weight as a complement.}\par

\section{Conclusion and Limitations}
In this work we experimentally explain the obstacle of recent methods of sparsely annotated object detection (SAOD). Hence we propose a Calibrated Teacher that adjusts the confidence of the teacher network's prediction adaptively to provide reliable pseudo labels. We further present the Focal IoU Weight as a complement to down-weight the false negative samples. Extensive methods show that our methods not only outperform the state-of-the-art counterparts, but also simplify the threshold tuning greatly. One limitation is that we only adopt a simple strategy for calibration, it is unclear that whether other advanced calibrators could help the performance. We leave it for future work.

\section{Acknowledgments}
This work is partially supported by the NSFC fund (61831014), in part by the Shenzhen Science and Technology Project under Grant (CJGJZD20200617102601004, ZDYBH201900000002).

\bibliography{aaai23}

\end{document}


\maketitle

Thanks for reading the supplementary materials. Here we will introduce more ablation study and plentiful qualitative comparisons.

\section{More Ablation Study}
\subsection{The Effect of Strong Augmentation}
In the teacher-student paradigm, we apply strong augmentation to the input of the teacher network. On the one hand, the strong augmentation benefits the diversity between the teacher and student networks. On the other hand, it increases the difficulty of the training of the student network in order to prevent it from over-fitting the false pseudo labels generated by the teacher network. However, data augmentation itself is also a common strategy to improve the performance. In order to explore how much the strong augmentation plays a role in the improvements we made in total, we further conduct the experiments in Easy, Hard and Extreme sets. \par

As shown in Tab. \ref{strongaug}, the strong augmentation itself could only improve the performance of the naive baseline by a little margin (0.1 $\sim$ 0.5 mAP). This is far from the improvement resulting from our framework, which is 1.1 $\sim$ 2.5 mAP. Therefore, our framework provides more than naive data enhancement. \par

\subsection{The Effect of Dynamic Threshold}

In this subsection, we evaluate the dynamic threshold in Split-1,2,3. As shown in Tab. \ref{DT}, the dynamic threshold outperforms the baseline in all settings. This proves that a gradually increasing threshold could benefit the training. However, it is still challenging to design a proper dynamic strategy. Therefore, the problem of tuning parameters has not been fundamentally solved. \par

\begin{table}[t] 
\resizebox{\columnwidth}{!}{
\begin{tabular}{@{}lccc@{}}
\toprule
Method & Easy & Hard & Extreme \\ \midrule
\begin{tabular}[c]{@{}l@{}}Baseline \\ \textit{w/o strong} aug\end{tabular} & 37.0 & 34.9 & 26.2 \\
\begin{tabular}[c]{@{}l@{}}Baseline \\ \textit{w/ strong} aug\end{tabular} & 37.1(+0.1) & 35.3(+0.4) & 26.7(+0.5) \\
Ours & \textbf{38.1(+1.1)} & \textbf{36.4(+1.5)} & \textbf{28.7(+2.5)} \\ \bottomrule
\end{tabular}
}
\caption{The effect of strong augmentation} 
\label{strongaug}
\end{table}

\begin{table}[t]
\centering
\resizebox{\columnwidth}{!}{
\begin{tabular}{lccccccccc}
\toprule
\multirow{2}{*}{Method} &  \multicolumn{3}{c}{Split-1} & \multicolumn{3}{c}{Split-2} & \multicolumn{3}{c}{Split-3} \\ \cmidrule[\cmidrulewidth](l){2-4} \cmidrule[\cmidrulewidth](l){5-7} \cmidrule[\cmidrulewidth](l){8-10}
 & 30\% & 50\% & 70\% & 30\% & 50\% & 70\% & 30\% & 50\% & 70\% \\ \midrule

Baseline & 39.3 & 37.5 & 34.1 & 39.1 & 37.4 & 35.5 & 39.2 & 37.6 & 34.3 \\ \midrule

DT & \textbf{39.7} & \textbf{38.5} & \textbf{35.5} & \textbf{39.5} & \textbf{38.2} & \textbf{36.9} & \textbf{39.4} & \textbf{38.1} & \textbf{35.7} \\

\bottomrule
\end{tabular}
}
\caption{The effect of dynamic threshold (DT)}
\label{DT}
\end{table}

\begin{table}[t]
\centering
\resizebox{0.75\columnwidth}{!}{
\begin{tabular}{@{}lccc@{}}
\toprule
Method & Easy & Hard & Extreme \\ \midrule
Co-mining & 35.4 & 31.8 & 23.0 \\
Ours & \textbf{35.5} & \textbf{33.5} & \textbf{26.5} \\ \bottomrule
\end{tabular}
}
\caption{Comparison with Co-mining in Easy, Hard and Extreme sets}
\label{90k}
\end{table}

\subsection{Comparison with Co-mining}

\whh{In this subsection, following the setting of Co-mining, we train our model for $90k$ iterations, decrease the learning rate at $60k$ and $80k$ iterations, and evaluate the model in Easy, Hard and Extreme sets. As shown in Tab. \ref{90k}, our methods still outperform Co-mining in all three sets. }

\subsection{The Comparison of ECE}

\begin{figure*}[t]
	\centering
	\includegraphics[width=0.98\textwidth]{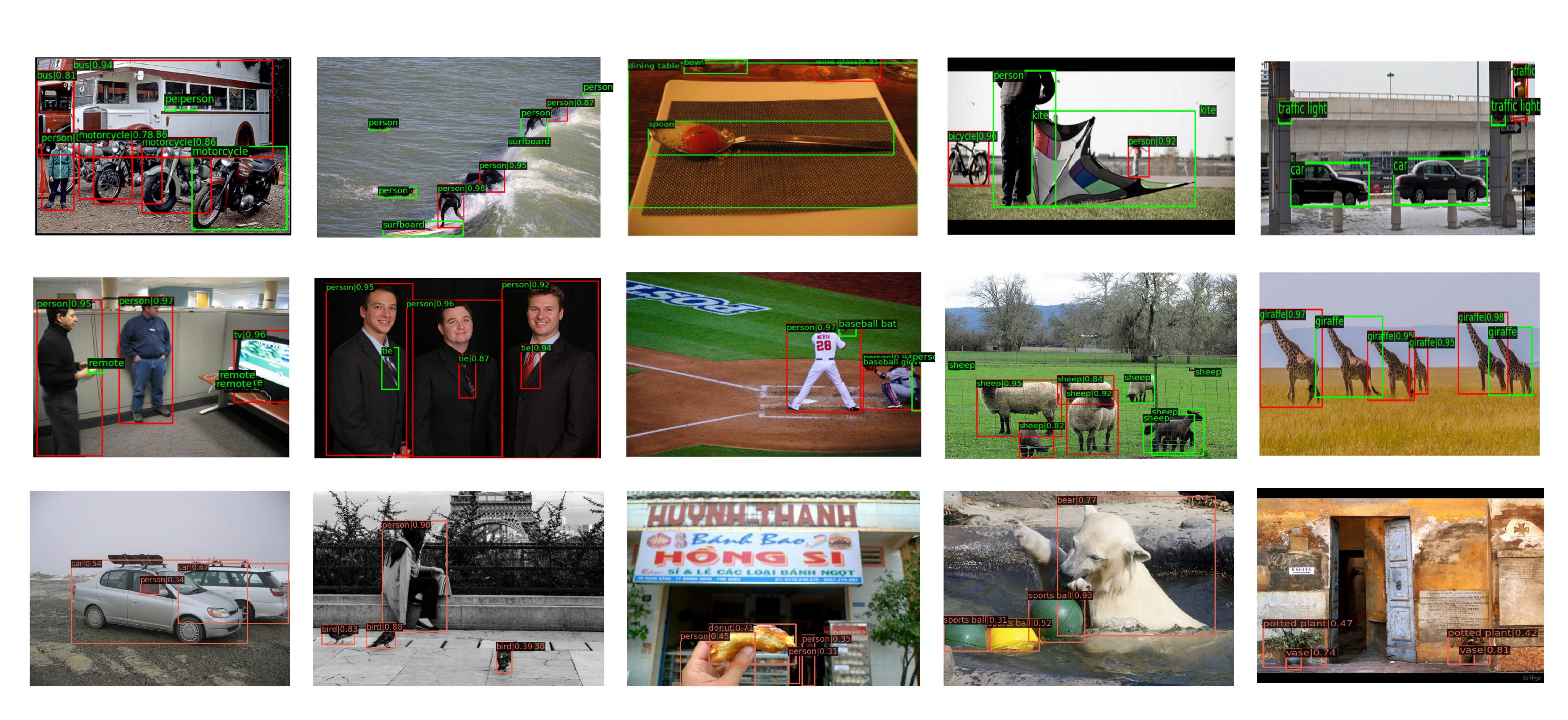}
	\caption{Qualitative visualization for the results of sparsely annotated object detection. The 1st and 2nd rows visualize the pseudo labels generated during training, where the green boxes correspond to the available ground truth, and the red boxes indicate the mined pseudo labels. The 3rd row shows the results of COCO validation set. }
	\label{fig:sup_visual}
\end{figure*}

\begin{figure}[t]
	\centering
	\includegraphics[width=0.95\columnwidth]{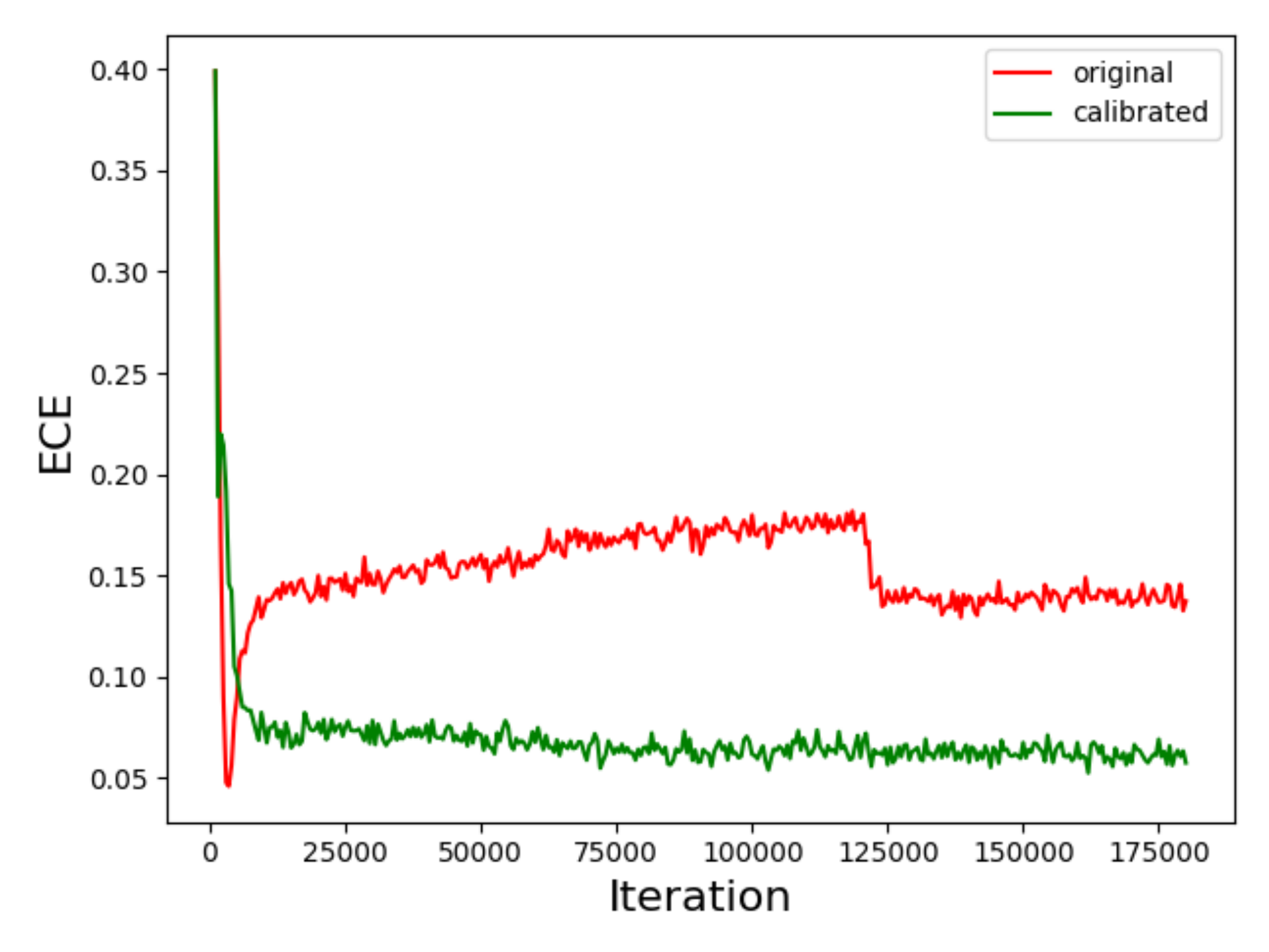}

	\caption{Comparison on the ECE during the whole training}
	\label{fig:supp_timely}
\end{figure}

\begin{figure}[t]
	\centering
	\includegraphics[width=0.95\columnwidth]{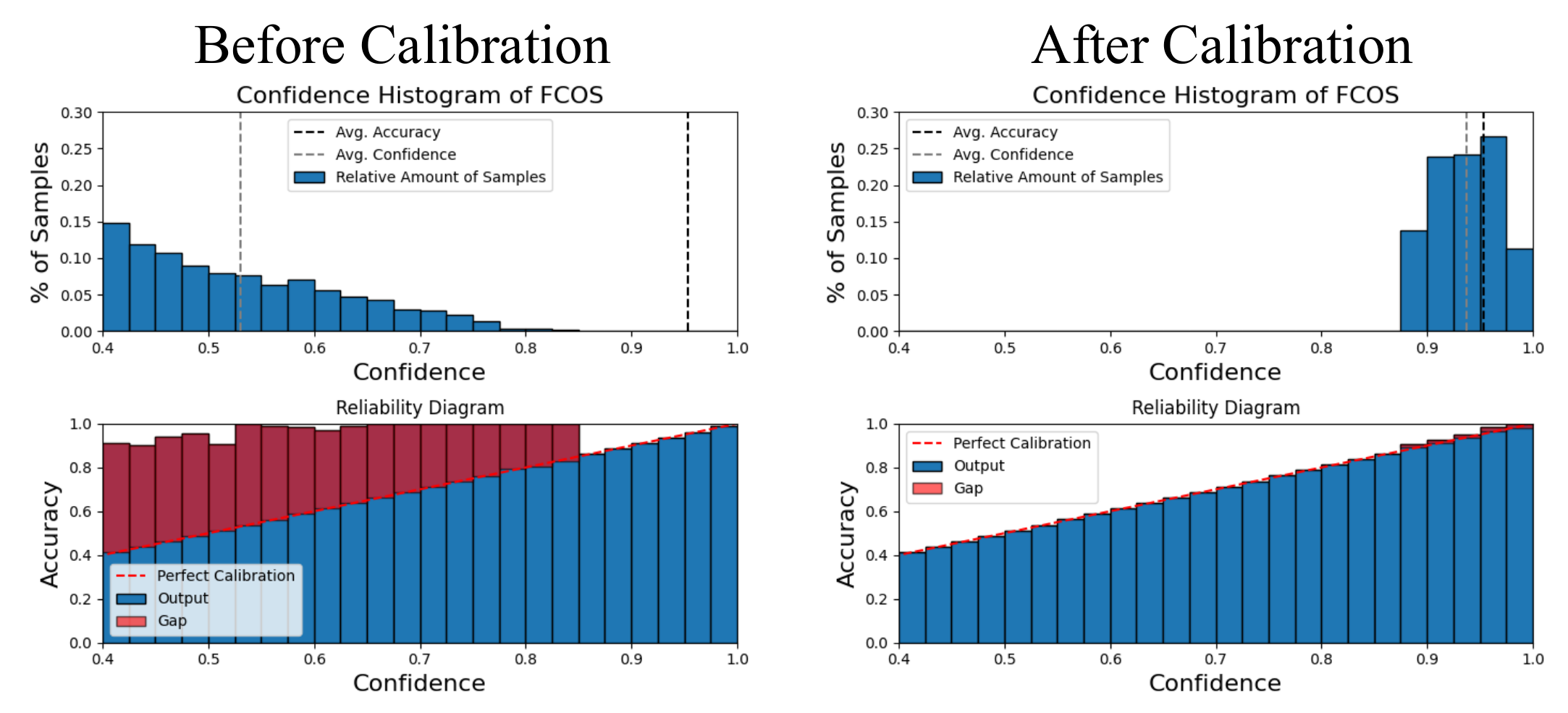}

	\caption{Comparison on the score distribution.}
	
	\label{fig:sup_ece}
\end{figure}

In this subsection we compare the Expected Calibration Error (ECE) of original confidence distribution and calibrated confidence distribution of RetinaNet in the whole training. As illustrated in Fig. \ref{fig:supp_timely}, the ECE drops significantly after calibration. It turns out that our calibrator works as expected. Therefore, our strategy of optimizing the calibrator (the design of IoU filter and queue) is effective in sparsely annotated object detection. \par

Besides we visualize the confidence distribution of FCOS (the distribution of RetinaNet and Faster R-CNN is illustrated in the paper). As shown in Fig. \ref{fig:sup_ece}, the calibrated confidence could keep consistent with its precision after calibration. Hence it proves that our strategy of confidence calibration is model-agnostic, and could be applied to nearly all the existing detectors conveniently. \par

\section{More Qualitative Results}

\subsection{Visualization of Pseudo Labels $\textbf{\&}$ Detection Results }
We show more qualitative results in Fig. \ref{fig:sup_visual}. As shown in the 1st row, our Calibrated Teacher is able to mine the missing instances in the dense scene (e.g. the motorcycle in the 1st column), the tiny instances (e.g. the persons in the 2nd and 4th columns, the traffic lights in the 5th column), the partially visible instances (the wine glass in the 3rd column), and blurry instances (the top-right person in the 2nd column), which are easy to be ignores during annotating. Then it could be found in the 2nd row that our method is able to provide pseudo labels with extremely high quality in the relatively simple scene. Hence it is possible to only annotate part of the instances to reduce the high labor and time costs. The 3rd row provide some results in the validation set. Our model is even able to detect the person behind the glass (the 1st column), the instances in the black-and-white picture (the 2nd column), the not salient person (the 3rd column), the partially visible ball (the 4th column) and the tiny vases (the 5th column). The qualitative results in such a challenging scene demonstrates the effectiveness of our method. \par